\documentclass[journal]{IEEEtran}
\usepackage{shtabularlines,array}
\usepackage{color}

\ifCLASSINFOpdf
  \usepackage[pdftex]{graphicx}
  \graphicspath{{../pdf/}{../jpeg/}}
  \DeclareGraphicsExtensions{.pdf,.jpeg,.png}
\else
  \usepackage[dvips]{graphicx}
  \graphicspath{{../eps/}}
  \DeclareGraphicsExtensions{.eps}
\fi
\usepackage{amsmath}
\begin{document}
%
\title{Effects of Image Degradations to CNN-based Image Classification}
%
%
%

\author{{Yanting Pei,
        Yaping Huang,
        Qi Zou,
        Hao Zang,
        Xingyuan Zhang,
        Song Wang}
\thanks{Y. Pei, Y. Huang, Q. Zou, H. Zang and X. Zhang are with the Beijing Key Laboratory of Traffic Data Analysis and Mining, Beijing Jiaotong University, Beijing, 100044, China (E-mail: \{yantingpei, yphuang, qzou, 16120451, 15112071\}@bjtu.edu.cn)}
\thanks{S. Wang is with the Department of Computer Science and Engineering, University of South Carolina,
Columbia, SC, USA (E-mail:songwang@cec.sc.edu)}}
\maketitle


\begin{abstract}
Just like many other topics in computer vision, image classification has achieved significant progress recently by using deep-learning neural networks, especially the Convolutional Neural Networks (CNN). Most of the existing works are focused on classifying very clear natural images, evidenced by the widely used image databases such as Caltech-256, PASCAL VOCs and ImageNet. However, in many real applications, the acquired images may contain certain degradations that lead to various kinds of blurring, noise, and distortions. One important and interesting problem is the effect of such degradations to the performance of CNN-based image classification.  More specifically, we wonder whether image-classification performance drops with each kind of degradation, whether this drop can be avoided by including degraded images into training, and whether existing computer vision algorithms that attempt to remove such degradations can help improve the image-classification performance. In this paper, we empirically study this problem for four kinds of degraded images -- hazy images, underwater images, motion-blurred images and fish-eye images. For this study, we synthesize a large number of such degraded images by applying respective physical models to the clear natural images and collect a new hazy image dataset from the Internet. We expect this work can draw more interests from the community to study the classification of degraded images.
\end{abstract}


%
\IEEEpeerreviewmaketitle

\section{Introduction}
%
%
%
%

Associating an input image with one of the priorly specified image class, \emph{image classification} is a fundamental and important problem in computer vision and artificial intelligence~\cite{yang2009linear,sanchez2011high}. While image classification has been studied in different applications for a long time, its performance is substantially improved in recent years by using supervised deep learning, e.g., Convolutional Neural Networks (CNN)~\cite{krizhevsky2012imagenet,simonyan2015very,he2016deep}, which unifies the feature extraction and classification into a single end-to-end network. For example, on ImageNet dataset,  a recent CNN-based image-classification method~\cite{he2016deep} achieved a top-5 accuracy of 96.4\%.

\begin{figure}[!t]
\centering
\includegraphics[width=3.5in]{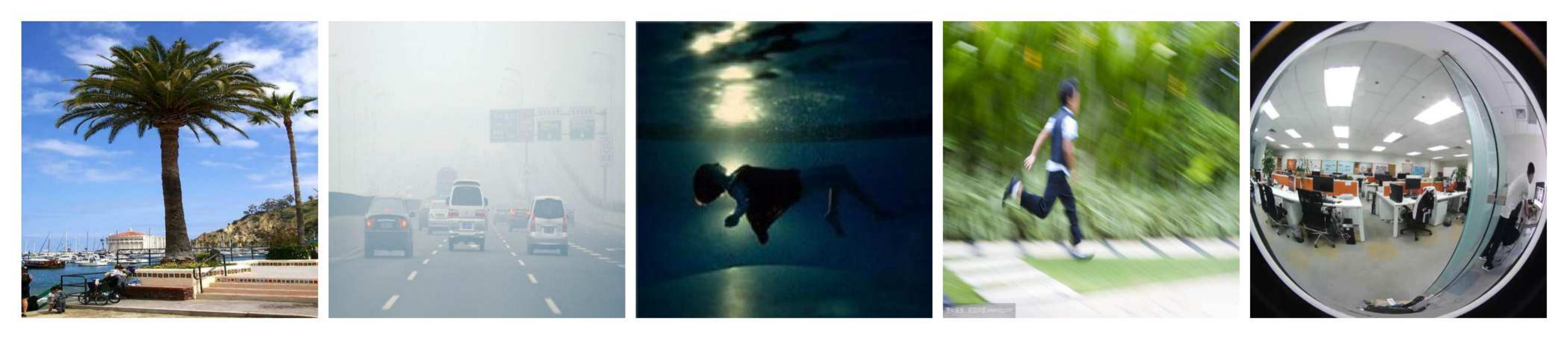}
\caption{Examples of clear and degraded images. From left to right are a clear natural image, a hazy image, an underwater image, a motion-blurred image and a fish-eye image, respectively.}
\label{fig:image-samples}
\end{figure}

However, most of these excellent image classification performances are achieved on clear natural images, such as the images in databases of Caltech-256~\cite{griffin2007caltech}, PASCAL VOCs~\cite{everingham2010pascal} and ImageNet~\cite{deng2009imagenet}. In many real applications, such as those related to autonomous driving, underwater robotics, video surveillance, and wearable cameras, the acquired images are not always clear. Instead, they suffer from various kinds of degradations. For example, images taken in the hazy weather, images take underwater by waterproof cameras, and image taken by moving cameras usually contain different levels of intensity blurs. Images taken by fish-eye cameras usually show spatial distortions. Some examples are shown in Fig.~\ref{fig:image-samples}. One important and interesting problem is whether the excellent classification performance obtained on clear natural images can be preserved on such degradation images by using the same deep learning techniques. In this paper, we empirically study this problem by constructing datasets of various kinds of degraded images and quantitatively evaluate and compare the CNN image classification models on these datasets.

More specifically, in this paper we select four kinds of degraded images -- hazy images, underwater images, motion-blurred images and fish-eye images -- for our empirical study. To quantify the classification performance under different level of degradations, we use their respective physical models to synthesize a large number of degraded images, as well as collecting real hazy images from Internet. We then implement the CNN model using AlexNet~\cite{krizhevsky2012imagenet} and VGGNet-16~\cite{simonyan2015very} on Caffe and use them for classifying images with different degradation levels. CNN-based model employs supervised learning and requires a set of images for training. To more comprehensively explore the effect of degradations to image classification, we study not only the training and testing on the images with the same level of degradation, but also the training and testing on images with different levels of degradations. 

The main contributions of this paper are threefolds. First, we conduct an empirical study of the effect of four kinds of typical image degradations to the CNN-based image classification. Second, we propose the use of respective physical models to construct synthetic images with different levels of degradations for quantitative evaluation. Third, we investigate whether existing degradation removal algorithms can benefit degraded image classification. Note that the goal of this paper is not the development of a new method to improve the classification performance on degraded images. Instead, we study whether the degraded image classification is a more challenging problem compared to the classification of clear natural images and whether its performance can be improved by selecting or pre-processing training/test data. We expect this work can help attract more interests from the community to further study the image classification of degraded images, including the development of new methods for improving the image-classification performance. 

The remainder of the paper is organized as follows. Section~\ref{sec:related} overviews the related work. Section~\ref{sec:proposed} introduces the construction of degraded images, the image-classification evaluation metric, and the CNN-based image classification model. Section~\ref{sec:experiment} reports the training and testing datasets, experimental results, and further analysis of the experimental results, followed by a brief conclusion in Section~\ref{sec:conclusion}.

\section{Related Work} \label{sec:related}

Just like many other topics in computer vision, the performance of \textbf{image classification} has been significantly improved by using deep learning techniques, especially the Convolution Neural Networks (CNN). In 2012, a CNN-based method~\cite{krizhevsky2012imagenet} achieved a top-5 classification accuracy of 83.6\% on ImageNet dataset in the ImageNet Large-Scale Visual Recognition Challenge 2012 (ILSVRC 2012). It is about 10\% higher than the traditional methods~\cite{sanchez2011high} that achieved a top-5 accuracy of 74.3\% on ImageNet dataset in ILSVRC 2011. Almost all the recent works and the state-of-the-art performance on image classification were achieved by CNN-based methods. For example, VGGNet-16 in~\cite{simonyan2015very} increased the network depth using an architecture with very small ($3\times 3$) convolution filters and achieved a top-1 accuracy of 75.2\% and a top-5 accuracy of 92.5\% on ImageNet dataset in ILSVRC 2014. Image classification accuracy in ILSVRC 2014 was then further improved~\cite{szegedy2015going} in 2015 by increasing the depth and width of the network. In~\cite{he2016deep}, residual learning was applied to solve the gradient disappearance problem and achieved a top-5 accuracy of 96.4\% on ImageNet dataset in ILSVRC 2015. In~\cite{hu2017squeeze}, an architectural unit was proposed based on the channel relationship,  which adaptively  recalibrates the channel-wise feature responses by explicitly modeling interdependencies between channels, resulting in a top-5 accuracy of 97.8\% on ImageNet dataset in ILSVRC 2017. In~\cite{Durand2017WILDCAT},  image regions for gaining spatial invariance are aligned and strongly localized features are learned, resulting in  95.0\%, 93.4\%, 94.4\% and 84.0\% classification accuracies on PASCAL VOC 2007, PASCAL VOC 2012, Scene-15 and MIT-67 datasets, respectively. Although these CNN-based methods have achieved excellent performance on image classification, most of them were only applied to the classification of clear natural images.

Degraded image-based recognition and classification have been studied in several recent works. In~\cite{karahan2016image}, only the influence of special degradations to face recognition was analyzed when using deep CNN-based approaches. In this paper, we investigate the problems of general degraded image classification, by covering hazy images, motion blurs, underwater blurs, and fish-eye distortions. Furthermore, In~\cite{karahan2016image}, the training data are always clear images while in this paper we will study whether the direct use of degraded images for training is beneficial or not. In~\cite{wang2016studying}, special degradations of low image resolution was studied in the applications of face identification, digit recognition and font recognition. In ~\cite{liu2017enhance}, a CNN-based method was proposed for improving the recognition performance of low-quality images and video by using pre-training, data augmentation, and other strategies. In this paper, we c onduct an empirical study to comprehensively understand the effects of various degradations to the performance of CNN-based image classification and investigate whether the use of degraded image in training and a pre-processing of degradation removal are helpful for image classification, which have important guiding significance to future work.
 

For \textbf{hazy images}, many models and algorithms were developed for removing the haze and restore the original clear image. He et al.~\cite{he2011single} presented a single-image haze-removal method using the dark channel prior. Zhu et al.~\cite{zhu2015fast} presented a single-image haze-removal algorithm using the color attenuation prior. Berman et al.~\cite{berman2016non} introduced a haze removal method based on the haze line. Cai et al.~\cite{cai2016dehazenet} adopted CNN-based deep architecture, whose layers are specially designed to embody the established priors in image dehazing and it is constructed by three convolution layers, a max-pooling, a Maxout unit and a BReLU activation function. Ren et al.~\cite{ren2016single} proposed a multi-scale deep neural network for haze removal, and the network consists of a coarse-scale net for a holistic transmission map and a fine-scale net for local refinement. Li et al.~\cite{li2017aod} designed an end-to-end network based on a re-formulated atmospheric scattering model, instead of estimating the transmission matrix and the atmospheric light separately. Recently, researchers also investigated haze removal from the images taken at nighttime hazy scenes. For example, Li et al.~\cite{li2015nighttime} developed a method to remove the nighttime haze with glow and multiple light colors. Zhang et al.~\cite{zhang2017fast} proposed a fast nighttime haze removal method using the maximum reflectance prior.
 
Different from the standard indoor and outdoor environments, the visible distance in many underwater conditions is only few meters. The \textbf{underwater images} taken by waterproof cameras, or other imaging facilities, are usually highly blurred and recognizing the objects from an underwater image is an important problem for both civil and military applications.  In \cite{oliver2010feature},  a traditional feature matching method using linear sparse coding is developed for underwater object recognition/detection. Jordt et al.~\cite{jordt2013refractive} proposed a system for computing camera path and 3D points from underwater images. Yau et al.~\cite{yau2013underwater} extended the existing works on physical refraction models by considering the dispersion of light, and derived new constraints on the model parameters for underwater camera calibration. Sheinin et al.~\cite{sheinin2016next} generalized the next best view concept of robot vision to scattering media and cooperative movable lighting for underwater navigation. Akkaynak et al.~\cite{akkaynak2017space} introduced the space of attenuation coefficients that can be used for many underwater computer vision tasks. Wang et al.~\cite{wang2017feeble} proposed a method for feeble object detection of underwater images through logical stochastic resonance with delay loop. Moller et al.~\cite{moller2017active} proposed a active learning method for the classification of species in underwater images from a fixed observatory. Rajeev et al.~\cite{rajeev2018improved} proposed a segmentation technique for underwater images based on K-means and local adaptive thresholding. Chen et al.~\cite{chen2018underwater} proposed a underwater object segmentation method based on optical features.

The motion of the camera and/or the captured objects usually introduces \textbf{motion blur} to the acquired images. Liu et al.~\cite{liu2008image} proposed a blurred image classification and analysis framework for detecting images containing blurred regions and recognizing the blur types for those regions without needing to perform blur kernel estimation and image deblurring. Golestaneh et al.~\cite{Golestaneh2017Spatially} proposed a spatially-varying blur detection method. Kalalembang et al.~\cite{Kalalembang2010DCT} presented a method of detecting unwanted motion blur effects. Gast et al.~\cite{Gast2016Parametric} proposed a parametric object motion model by combining with a segmentation mask to exploit localized, non-uniform motion blur. Lin et al.~\cite{lin2011motion} addressed the problem of matting motion blurred objects from a single image. Besides, effective and efficient deblurring of such degraded images have become an important research topic in the past decades. In~\cite{kim2016accurate}, an end-to-end algorithm was developed to reconstruct motion-blur-free images. In~\cite{sun2015learning}, the motion flow was estimated from a single degraded image and then compensation was made to remove motion blur. Fan et al.~\cite{fan2017image} proposed a new blur classification model using convolutional neural network. 

\textbf{Fish-eye images} can provide wide-angle view of a scene, but introduce distortions to the covered scene and objects. Kannala et al.~\cite{kannala2006generic} proposed a generic camera model for both the conventional and wide-angle lens cameras, as well as developing a calibration method for estimating the parameters of the model. Fu et al.~\cite{fu2012forgery} discussed how to explicitly employ the distortion cues to detect the forgery object in fish-eye images. Hughes et al.~\cite{Hughes2010Equidistant} proposed a method to estimate the intrinsic and extrinsic parameters of fish-eye cameras. Wei et al.~\cite{Wei2012Fisheye} proposed a fish-eye video correction method. Ying et al.~\cite{ying2006fisheye} presented a method to calibrate fish-eye lenses. Li et al.~\cite{li2018fisheye} proposed a new fish-eye image rectification method that combines the physical spherical model and the digital distortion model. Krams et al.~\cite{krams2017people} addressed the problem of people detection in top-view fish-eye imaging.  Baek et al.~\cite{baek2018real} proposed a method for real-time detection, tracking, and classification of moving and stationary objects using multiple fish-eye images.

Different from these prior works on hazy, underwater, motion-blur and fish-eye images, in this paper, we conduct a comprehensive empirical study to quantify the effects of these four kinds of degradations to image classification. Some of these prior works investigated the removal of degradations. Later in this paper, we will study whether the removal of degradations using these methods can help the CNN-based image classification or not.

\section{Proposed Method} \label{sec:proposed}
 
In this section, we first discuss the construction of the images with different levels of degradations, followed by CNN-based image classification models and performance metric.
 
\subsection{Synthesis of Degraded Images}

There are two difficulties in quantitatively and comprehensively evaluating the CNN-based degraded image classification. First, it is very difficult to collect a large number of real degraded images with the desired class information for training the CNN models. Second, the degradation level of the collected real images are usually unknown and therefore, the use of such images could not quantify the effect of degradation levels to the image classification performance. To address these problems, we propose to first synthesize degraded images for large-scale CNN-based training and testing. 

There are many available image datasets, such as Caltech-256, PASCAL VOCs and ImageNet, consisting of different classes of clear natural images, that have been widely used for evaluating image classification models. We can select one such dataset, take each image in this dataset as the original image without any degradation, and then synthesize its degraded versions by using available physical models. In this data synthesis, we can control the level of the added degradations. By using these synthesized degraded images for training and testing the CNN models, we can systematically study the effect of different levels of degradations to the performance of image classification. 

We synthesize hazy images by~\cite{zhang2017fast}
\begin{eqnarray*}
\emph{\textbf{I}}(\emph{\textbf{x}})=t(\emph{\textbf{x}})\cdot\emph{\textbf{J}}(\emph{\textbf{x}})+[1-t(\emph{\textbf{x}})]\cdot\emph{\textbf{A}},
\end{eqnarray*}
where $\emph{\textbf{x}}$ is the pixel coordinates, $\emph{\textbf{I}}$ is the synthesized hazy image, $\emph{\textbf{J}}$ is the original image and $\emph{\textbf{A}}$ is the global atmospheric light.
The scene transmission $t(\emph{\textbf{x}}$) is distance-dependent:
\begin{eqnarray*}
t(\emph{\textbf{x}})=e^{-\beta d(\emph{\textbf{x}})},
\end{eqnarray*}
where $\beta$ is the atmospheric scattering coefficient and $d(\emph{\textbf{x}})$ is the normalized distance of the scene at pixel $\emph{\textbf{x}}$. We get the depth map $d(\emph{\textbf{x}})$ by following~\cite{Depth2015CVPR}. We can control the degradation levels of the synthesized hazy images by varying $\beta$.
 
\begin{figure*}[!t]
\centering
\includegraphics[width=6.8in]{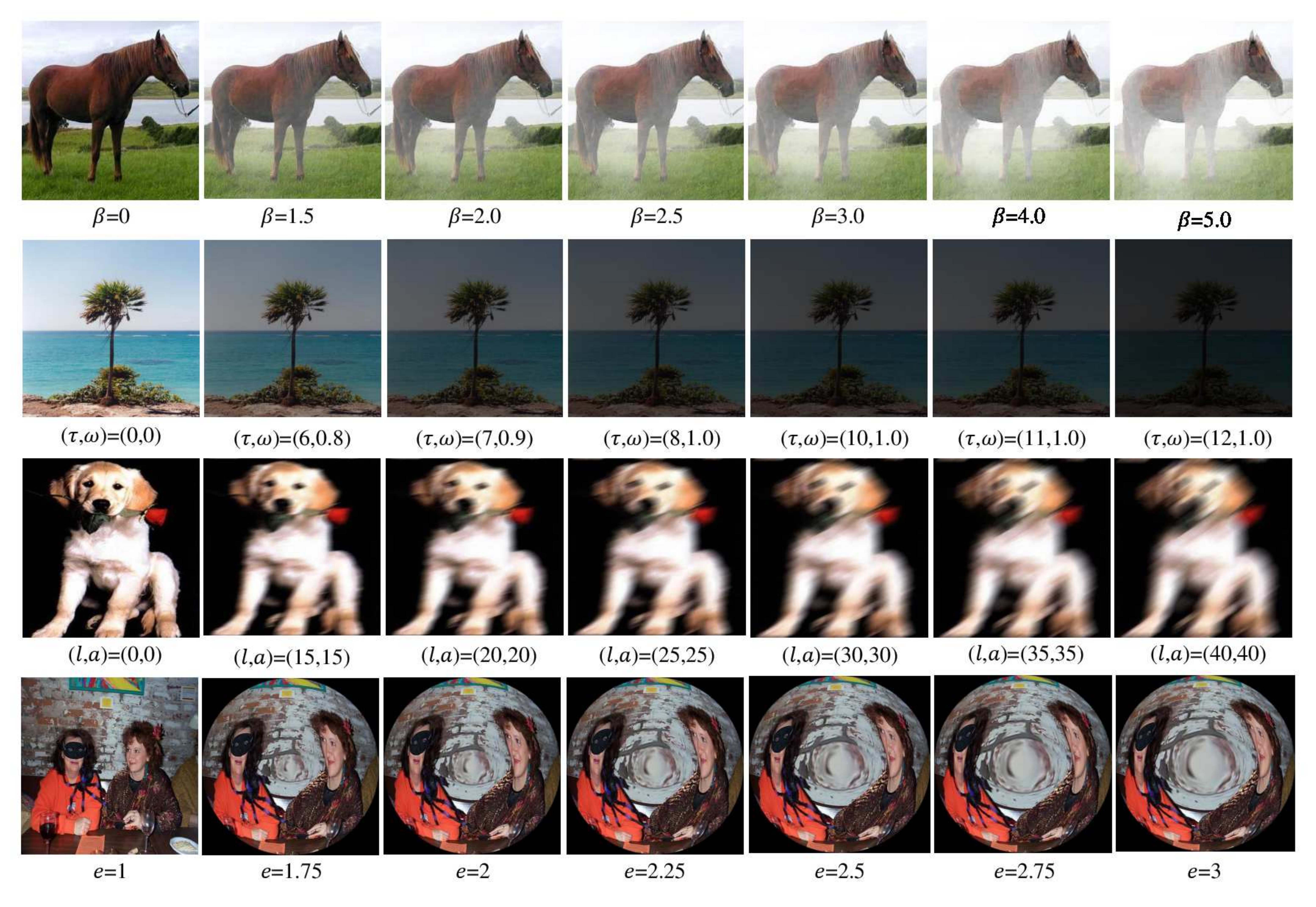}
\caption{Examples of the synthesized degraded images. From top to the bottom rows are hazy images, underwater images, motion-blurred images and fish-eye images. The degradation levels are indicated below each image. The first column shows the original images.}
\label{fig:synthetic-samples}
\end{figure*}

We synthesize underwater images by convolving the input image with the simplified version of Dolin’s PSF model~\cite{dolin2006theory,oliver2010feature}
\begin{eqnarray*}
\begin{split}
&G(\theta_b,\tau_b)=\frac{\delta(\theta_q)}{\pi \theta_q}\exp(-\tau_b)+0.525\frac{\tau_b}{\theta_q}\\ 
&\quad \quad \cdot \exp(-2.6{\theta_q}^{0.7}-\tau_b)+\frac{\beta_2^2}{2\pi}[2-(1+\tau_b)\exp(-\tau_b)]\\ 
&\quad \quad \cdot \exp[-\beta_1(\beta_2\theta_q)^\frac{1}{3}-(\beta_2 \theta_q)^2+\beta_3],
\end{split}
\end{eqnarray*}
where $\theta_q$ is the scattering angle, $\tau_b=\tau \omega$, with $\tau$ being the optical depth and $\omega$ being the single scattering albedo. We can control the degradation levels of the synthesized underwater images by varying $\tau$ and $\omega$.

We synthesize motion-blurred images by following~\cite{sun2015learning} and a motion-blurred image can be modeled as follows:
\begin{eqnarray*}
\emph{\textbf{G}}(\emph{\textbf{x}})=\emph{\textbf{H}}(l,a)*\emph{\textbf{F}}(\emph{\textbf{x}})+\emph{\textbf{N}}(\emph{\textbf{x}}),
\end{eqnarray*}
where $\emph{\textbf{G}}(\emph{\textbf{x}})$ is a motion-blurred image， $\emph{\textbf{H}}(l,a)$ is the motion blur kernel, and $\emph{\textbf{F}}(\emph{\textbf{x}})$ is the clear natural image. * is the convolution operator and $\emph{\textbf{N}}(\emph{\textbf{x}})$ is an additive noise. $\emph{\textbf{x}}$ is the pixel coordinates. The degradation level is controlled by varying the parameters $l$ and $a$, with $l$ being the length of the blur kernel and $a$ being the counterclockwise rotation angle of the object.

We synthesize fish-eye images by following~\cite{kannala2006generic}, where the pixel coordinates $(u,v)$ of the fish-eye image is computed from 
\begin{eqnarray*}
\left(\begin{matrix}
u\\
v
\end{matrix}\right)
=\left[
\begin{matrix}
n_u &0\\
0 &n_v
\end{matrix}
\right]
\left(
\begin{matrix}
x_d\\
y_d
\end{matrix}
\right)
+\left(
\begin{matrix}
u_0\\
v_0
\end{matrix}
\right),
\end{eqnarray*}
where $(x_d,y_d)$ is the distorted coordinates and $x_d=Rcos\theta$ and $y_d=Rsin\theta$. $R=min(|\frac{1}{sin\theta}|,|\frac{1}{cos\theta}|)r^e$ with $r=\sqrt{(x^2+y^2)}$ and $\theta=atan(\frac{y}{x})$. $(x,y)$ is the coordinates of the original image. $(u_0,v_0)$ is the principal point and $n_u$  and $n_v$ are the number of pixels per unit distance along horizontal and vertical directions, respectively. The degradation level is controlled by varying the exponent $e$.

Figure~\ref{fig:synthetic-samples} shows the examples of the synthesized images of different degradation levels. 

\subsection{CNN-based Image Classifiers and Evaluation Metric}
 
\begin{figure}[!t]
\centering
\includegraphics[width=3.3in]{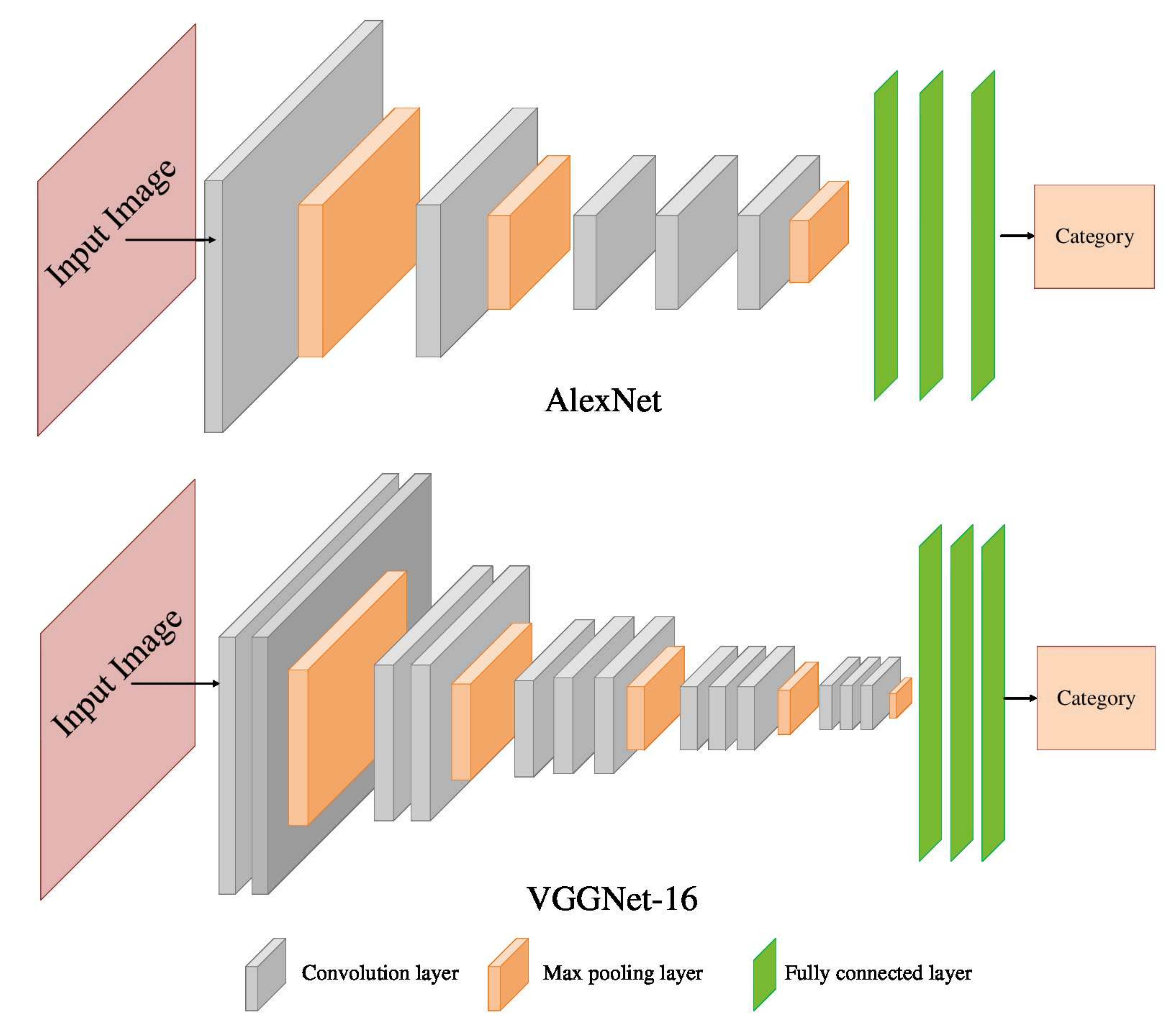}
\caption{The CNN network architecture used in this paper. The top subfigure is AlexNet with 8 weight layers and the bottom subfigure is VGGNet-16 with 16 weight layers.}
\label{fig:CNN}
\end{figure}

In this paper, we use AlexNet and VGGNet-16 on Caffe to implement the CNN-based image classification model. The architecture of AlexNet and VGGNet-16 are summarized in Fig.~\ref{fig:CNN}. where the convolution layers are shown in gray, the max-pooling layers are shown in orange, and the fully-connected layers are shown in light green.

The AlexNet~\cite{krizhevsky2012imagenet} has 8 weight layers (5 convolutional layers and 3 fully-connected layers). The first convolutional layer has 96 kernels of size $11\times11\times3$ and filters a $224\times224\times3$ input image with stride 2. The second convolutional layer filters the output of the first convolutional layer with 256 kernels of size $5\times5\times48$. The third convolutional layer has 384 kernels of size $3\times3\times256$ connected to the outputs of the second convolutional layer. The fourth convolutional layer has 384 kernels of size $3\times3\times192$, and the fifth convolutional layer has 256 kernels of size $3\times3\times192$. Each fully-connected layer has 4,096 channels, except for the last fully-connected layer that has $C$ channels ($C$ is the number of classes). 

The VGGNet-16~\cite{simonyan2015very} has 16 weight layers (13 convolution layers and 3 fully-connected layers). It has very small $3\times3$ receptive fields with stride 1 throughout the whole net. The number of channels is small and it starts from 64 in the first layer and increases after each max-pooling layer by a factor of 2, until it reaches 512. The input is also a fixed-size $224\times224$ RGB image during training. Spatial pooling is carried out by a max-pooling layer, which follows the convolution layers. Max-pooling is performed over a $2\times2$ pixel window with stride 2. The first two fully-connected layers have 4,096 channels each and the third has 257 channels (one for each class). The final layer is the soft-max layer.
 
As detailed later, we use Caltech-256 dataset to synthesize the images with different degradation levels. We divide the synthesized image data for training and testing, and use the classification accuracy on the testing set as the performance evaluation metric. Specifically, the accuracy is defined by $\frac{R}{N}$, where $R$ is the number of correctly recognized images in the testing set and $N$ is the total number of testing images.

\subsection{Training and Testing Datasets}

For a comprehensive analysis of the effects of image degradations to CNN-based image classification, we vary the selection of training and testing image datasets for the CNN classifiers and design multiple experiments in our empirical study.

\begin{itemize}
\item For each kind of degradation, train and test CNN classifiers using image data of the different gradation levels.
\item For each kind of degradation, train and test CNN classifiers using image data of the same gradation level.
\item For each kind of degradation, train CNN classifiers by combining images of different degradation levels and test on images of each degradation level.
\end{itemize}
	
Finally, we collect a set of real hazy images from Internet for CNN training and/or testing to support our findings drawn from study on the synthetic data, although the exact degradation levels of these 
real images are unknown. 

\section{Experiments}\label{sec:experiment}

In this section, we first describe the datasets and experiment setup. After that, we report the experiment results on different datasets and visualize sample features extracted at each hidden layer to further analyze the experiment results. Finally, we conduct experiments to check whether a degradation-removal preprocessing step can improve the CNN-based image classification accuracy.

\subsection{Datasets and Experiment Setup}

We synthesize degraded images using all the images in Caltech-256 dataset~\cite{griffin2007caltech}, which has been widely used for evaluating image classification algorithms. This dataset contains 30,607 images from 257 classes (256 object categories and a clutter class). For each level of each kind of degradation, we synthesize 30,607 images by applying the respective model to each of the images in Caltech-256. In the original Caltech-256, we follow~\cite{simonyan2015very} to select 60 images as training images per class, and the rest are used for testing. Among the training images, 20\% per class are used as a validation set. We follow this to strategy split the synthesized image data: an image is in training set if it is synthesized from an image in the training set and in testing set otherwise. This way,  for each level of each kind of degradation, we have a training set of $60\times 257=15,420$ images (60 per class) and a testing set of  $30,607-15,420=15,187$ images. 

In our experiment, for each kind of degraded images, we select seven different levels of degradations. More specifically, for hazy images, we set the parameter $\beta \in \{0, 1.5, 2.0, 2.5, 3.0, 4.0, 5.0\}$. For underwater images, we set parameters $(\tau,\omega) \in \{ (0,0), (6, 0.8), (7, 0.9), (8, 1.0), (10, 1.0), (11, 1.0), (12, 1.0)\}$. For motion-blurred images, we set $(l,a)\in \{(0,0), (15, 15), (20, 20), (25, 25), (30, 30), (35, 35), (40, 40)\}$. For fish-eye images, we set parameter $e\in\{1, 1.75, 2, 2.25, 2.5, 2.75, 3\}$. For each kind of the degradations, the first level, i.e., the corresponding parameters are all zero, except fish-eye images is one, corresponds to the level without any degradation, which is equivalent to use the original images in Caltech-256 for experiments. 
 
\begin{figure}[!t]
\centering
\includegraphics[width=3.5in]{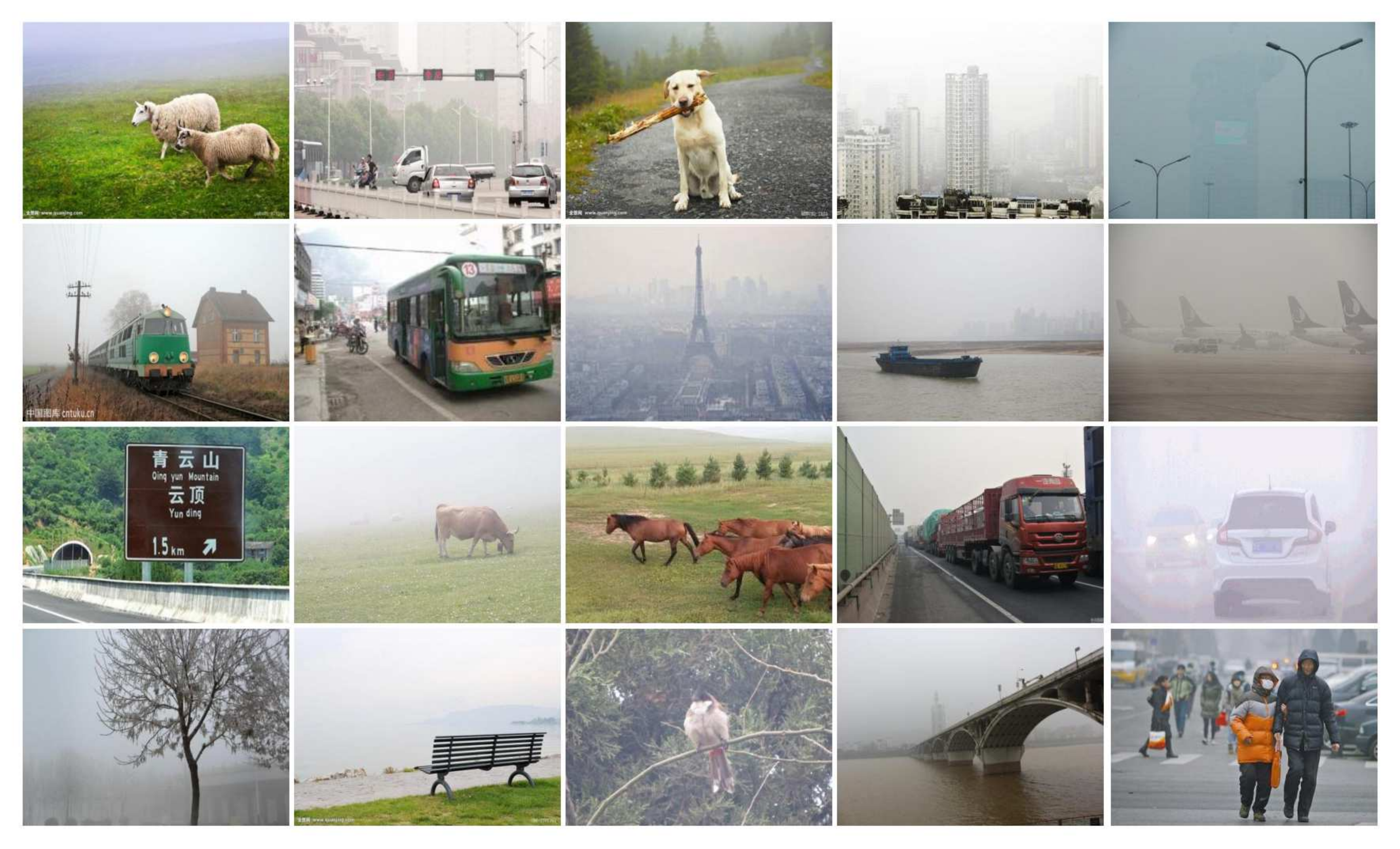}
\caption{Sample hazy images in our new Haze-20 dataset.}
\label{fig:haze-20}
\end{figure}

While we can construct synthetic degradation images by well-acknowledged physical models, real image degradations can be much more complicated and experiments on real images are still crucial. Given the relative wide availability of hazy images, we collect a new dataset of real hazy images from the Internet. This new dataset contains 4,610 images from 20 classes and we name it as \emph{Haze-20}. These 20 image classes are bird, boat, bridge, building, bus, car, chair, cow, dog, horse, people, plane, sheep, sign, street-lamp, tower, traffic-light, train, tree and truck, respectively. The number of images per class varies from 204 to 279, as given in Table~\ref{table:haze}. Some examples (one image of each class) in Haze-20 are shown in Fig.~\ref{fig:haze-20}. For the collected real hazy images in Haze-20, we select 100 images from each class as training images, and the rest are used for testing. Among the training images, 20\% per class are used as a validation set. So, we have a training set of $100 \times 20 = 2,000$ images and a testing set of $4,610-2,000 = 2,610$ images.

One important experiment in our empirical study is to train on clear images and test on degradation images. This is not an problem for our synthetic data since their underlying degradation-free clear images are available, i.e., the original Caltech-256 data. For collected real images in Haze-20, we do not have their underlying clear images. To address this issue, we collect a new \emph{HazeClear-20} image dataset from Internet, which consists of haze-free images that fall in the same 20 classes as in Haze-20. HazeClear-20 consists of 3,000 images, with 150 images per class.  For HazeClear-20 dataset, we also select 100 images from each class as training images, and the rest are used for testing. Among the training images, 20\% per class are used as a validation set. So, we have a training set of $100 \times 20 = 2,000$ images and a testing set of $50\times 20 = 1,000$ images.

\begin{table*}[htbp]
\renewcommand\arraystretch{1.5}
\centering
\setlength{\tabcolsep}{3mm}{
\caption{Number of images for each class in the new Haze-20 dataset.}
\label{table:haze}
\begin{tabular}{!{\shvline[2pt]}c!{\shvline[1pt]}c|c|c|c|c|c|c|c|c|c!{\shvline[2pt]}}
\shhline[2pt]
Haze-20 &bird &boat &bridge &build. &bus &car &chair &cow &dog &horse\\
\hline
Number &231 &236 &233 &251 &222 &256 &213 &227 &244 &237\\
\shhline[1pt]
Haze-20 &people &plane &sheep &sign &lamp & tower &light &train &tree &truck\\
\hline
Number &279 &235 &204 &221 &216 &230 &206 &207 &239 &223\\
\shhline[2pt]
\end{tabular}
}
\end{table*}

We implement AlexNet and VGGNet-16 on Caffe in this paper. The CNN architectures are pre-trained on ImageNet dataset that consists of 1,000 classes with 1.2 million training images. We use the pre-trained model for image classification by fine-turning on our training images. We change the number of channels in the last fully connected layer from 1,000 to $C$, where $C$ is the number of classes in our datasets. Note that in this paper we use AlexNet and VGGNet-16 for their simplicity -- it does not prevent from using other network structures, such as ResNet.

\subsection{Results on Synthetic Images -- Individual Degradation Levels}

For each kind of synthesized degraded images, we exhaustively try the training using the training set of one level and testing on the testing data of the same or another level. The image classification accuracies by using VGGNet-16 are summarized in the four subtables in Table~\ref{tab:accuracy-vggnet}, separated by thick lines. For each of the four subtables, the first column indicates the degradation level of the training set and the first row  indicates the degradation level of the testing set. For example, the accuracy corresponding to the row $\beta=1.5$ and the column $\beta=4.0$ is 65.0\%. This indicates the classification accuracy on the testing set at haze-degradation level $\beta=4.0$ is 65.0\% when using the VGGNet-16 classifiers trained using the training set at the haze-degradation level $\beta=1.5$. In these subtables, the accuracies along the diagonal are achieved by training and testing on the image data of the same degradation level. The accuracies at non-diagonal elements are achieved by training and testing on the image data of the different degradation levels. In these four subtables, we highlight the maximum accuracy in each column. 

\begin{table*}[htbp]%
\renewcommand\arraystretch{1.46}
\setbox0\hbox{\verb/\documentclass/}%
\caption{The classification accuracy (\%) of hazy images (upper-left subtable), underwater images (upper-right subtable), motion-blurred images (lower-left subtable) and fish-eye images (lower-right subtable) by using VGGNet-16 with different levels of degradation, respectively.}\label{tab:accuracy-vggnet}
\centering
\setlength{\tabcolsep}{1.2mm}{
\begin{tabular}{!{\shvline[2pt]}c!{\shvline[1pt]}c|c|c|c|c|c|c!{\shvline[2pt]}c!{\shvline[1pt]}c|c|c|c|c|c|c!{\shvline[2pt]}}
\shhline[2pt]
$\beta$ &\emph{0} &\emph{1.5} &\emph{2.0} &\emph{2.5} &\emph{3.0} &\emph{4.0} &\emph{5.0} &$(\tau,\omega)$ &\emph{(0,0)} &\emph{(6,0.8)} &\emph{(7,0.9)} &\emph{(8,1.0)} &\emph{(10,1.0)} &\emph{(11,1.0)} &\emph{(12,1.0)}\\
\shhline[1pt]
\emph{0} &\textbf{81.0} &74.6 &68.7 &61.3 &54.0 &41.8 &33.0 &\emph{(0,0)} &\textbf{81.0} &79.3 &75.3 &68.2 &57.4 &49.0  &40.3\\
\hline
\emph{1.5} &79.2 &\textbf{78.6} &77.6 &75.9 &73.5 &65.0 &55.1 &\emph{(6,0.8)} &80.0 &\textbf{80.4} &78.6 &75.8 &70.1 &66.3 &61.4\\
\hline
\emph{2.0} &78.5 &78.3 &\textbf{77.7} &76.5 &74.7 &68.6 &60.3 &\emph{(7,0.9)} &78.3 &79.5 &\textbf{79.8} &78.1 &74.6 &72.0 &68.2\\
\hline
\emph{2.5} &78.0 &78.1 &77.6 &\textbf{76.6} &75.2 &70.6 &63.2 &\emph{(8,1.0)} &76.7 &78.5 &79.5 &\textbf{78.9} &76.9 &74.9 &72.7\\
\hline
\emph{3.0} &76.7 &77.4 &77.0 &76.5 &\textbf{75.3} &71.4 &65.4 &\emph{(10,1.0)} &74.9 &76.5 &78.5 &78.5 &77.2 &76.1 &74.4\\
\hline
\emph{4.0} &75.2 &76.2 &76.1 &75.7 &75.1 &\textbf{72.3} &\textbf{68.1} &\emph{(11,1.0)} &74.1 &75.4 &77.7 &78.0 &\textbf{77.4} &\textbf{76.5} &74.9\\
\hline
\emph{5.0} &72.9 &74.4 &74.2 &73.9 &73.5 &71.2 &67.8 &\emph{(12,1.0)} &73.3 &74.0 &76.7 &77.2 &76.8 &76.2 &\textbf{74.9}\\
\shhline[2pt]
$(l,a)$ &\emph{(0,0)} &\emph{(15,15)} &\emph{(20,20)} &\emph{(25,25)} &\emph{(30,30)} &\emph{(35,35)} &\emph{(40,40)} &$e$ &\emph{1} &\emph{1.75} &\emph{2} &\emph{2.25} &\emph{2.5} &\emph{2.75} &\emph{3}\\
\shhline[1pt]
\emph{(0,0)} &\textbf{81.0} &45.1 &31.2 &22.6 &16.6 &13.3  &11.2 &\emph{1} &\textbf{81.0} &25.4 &17.6 &12.4 &9.4 &7.0 &5.1\\
\hline
\emph{(15,15)} &70.8 &\textbf{72.6} &69.0 &59.7 &45.3 &32.3 &23.4 &\emph{1.75} &71.7 &\textbf{73.7} &70.5 &65.9 &60.3 &53.7 &47.0\\
\hline
\emph{(20,20)} &67.1 &71.5 &\textbf{69.9} &65.9 &57.5 &44.7 &33.1 &\emph{2} &69.5 &73.2 &\textbf{71.0} &68.2 &64.6 &60.1 &55.0\\
\hline
\emph{(25,25)} &60.0 &67.9 &68.2 &\textbf{66.9} &63.9 &57.0 &47.3 &\emph{2.25} &65.5 &72.6 &71.0 &\textbf{69.2} &66.3 &62.9 &58.9\\
\hline
\emph{(30,30)} &55.4 &62.8 &64.5 &65.0 &\textbf{64.2} &61.0 &55.5 &\emph{2.5} &62.0 &71.1 &70.4 &68.9 &\textbf{66.9} &64.4 &61.6\\
\hline
\emph{(35,35)} &47.2 &55.4 &58.7 &61.0 &62.3 &\textbf{61.6} &58.5 &\emph{2.75} &59.6 &69.1 &68.9 &67.7 &66.0 &64.3 &62\\
\hline
\emph{(40,40)} &44.5 &47.3 &50.6 &54.9 &58.3 &59.8 &\textbf{59.0} &\emph{3} &56.4 &67.1 &67.2 &66.8 &65.6 &\textbf{64.5} &\textbf{62.7}\\
\shhline[2pt]
\end{tabular}%
}
\end{table*}
 
From the results reported in these four subtables, we can see that, the maximum values in each column are usually located along the diagonal of each subtable. This indicates that, to achieve the best possible accuracy in classifying the images with certain level of degradations, we need to collect the training images with the same kind of degradation and with the same or similar degradation levels. If the degradation levels of the training images and test images have large gaps, the testing accuracy can be very low. For example, if we use clear images ($\beta=0$) to train CNN classifiers, and then apply them to classify images with hazy level of $\beta=5.0$, the accuracy will drop significantly from $81.0\%$ (training and testing both on clear images) or $67.8\%$ (training and testing both on $\beta=5.0$ hazy images) to $33.0\%$. 
 
By examining the change of accuracy values along the diagonal of each subtable, we can see that they also decrease from the top-left element to the bottom-right element. This indicates that, even we train and test the CNN using the images of the same degradation level, the classification performance still drops when the degradation level increases. For example, in the lower-left subtable of Table~\ref{tab:accuracy-vggnet}, the classification accuracy drops to $59.0\%$ when we train and test both on $(l,a)=(40, 40)$-level motion-blurred images, while the accuracy is $81.0\%$ when both training and testing images have no degradation. This may be caused by the partial loss of discriminative image information in the image degradations. 

We conduct the same experiments using AlexNet and results are shown in the four subtables in Table~\ref{tab:accuracy-alexnet}. These results are largely consistent to the results shown in Table~\ref{tab:accuracy-vggnet}, e.g., in each subtable, the diagonal elements are usually lager than the non-diagonal ones, and the values along the diagonal drop from the top-left element to the bottom-right element.  We also find that, in general, VGGNet-16 produces higher classification accuracy than AlexNet when using the same training set. This is not surprising since VGGNet-16 is a deeper network.
For each subtable in Table~\ref{tab:accuracy-alexnet} and Table~\ref{tab:accuracy-vggnet}, we compute the relative accuracy drop by computing the difference between their top-left and bottom-right elements.
We can find that the accuracy drop in using VGGNet-16 is less than the drop in using AlexNet for three out four kinds of degradations  -- 13.2\%, 6.1\% and 18.3\%  (VGGNet-16)  v.s. 19.6\%, 14.3\% and 25.4\% (AlexNet) for hazy, underwater and fish-eye images, respectively.  Only for motion-blurred images, the accuracy drop (22.0\%) in using VGGNet-16 is a little more than the drop (20.3\%) in using AlexNet. From this perspective,
VGGNet-16 also outperforms AlexNet.

\begin{table*}[htbp]%
\renewcommand\arraystretch{1.46}
\setbox0\hbox{\verb/\documentclass/}%
\caption{The classification accuracy (\%) of hazy images (upper-left subtable), underwater images (upper-right subtable), motion-blurred images (lower-left subtable) and fish-eye images (lower-right subtable) by using AlexNet with different levels of degradation, respectively.}
\label{tab:accuracy-alexnet}
\centering
\setlength{\tabcolsep}{1.3mm}{
\begin{tabular}{!{\shvline[2pt]}c!{\shvline[1pt]}c|c|c|c|c|c|c!{\shvline[2pt]}c!{\shvline[1pt]}c|c|c|c|c|c|c!{\shvline[2pt]}}
\shhline[2pt]
$\beta$ &\emph{0} &\emph{1.5} &\emph{2.0} &\emph{2.5} &\emph{3.0} &\emph{4.0} &\emph{5.0} &$(\tau,\omega)$ &\emph{(0,0)} &\emph{(6,0.8)} &\emph{(7,0.9)} &\emph{(8,1.0)} &\emph{(10,1.0)} &\emph{(11,1.0)} &\emph{(12,1.0)}\\
\shhline[1pt]
\emph{0} &\textbf{71.7} &61.6 &53.9 &46.2 &39.1 &28.3 &21.7 &\emph{(0,0)} &\textbf{71.7} &68.0 &61.3 &47.8 &31.4 &23.7 &16.3\\
\hline
\emph{1.5} &68.0 &\textbf{68.5} &66.4 &63.9 &60.2 &50.0 &40.3 &\emph{(6,0.8)} &68.7 &\textbf{70.0} &66.0 &58.8 &44.7 &36.4 &27.9\\
\hline
\emph{2.0} &66.9 &67.9 &\textbf{66.5} &64.5 &61.7 &53.4 &44.4 &\emph{(7,0.9)} &65.4 &67.9 &\textbf{67.9} &64.2 &57.7 &52.5 &46.0\\
\hline
\emph{2.5} &65.7 &67.1 &66.1 &\textbf{64.6} &62.2 &55.8 &47.3 &\emph{(8,1.0)} &59.7 &65.2 &66.9 &\textbf{65.6} &62.0 &59.2 &55.4\\
\hline
\emph{3.0} &64.0 &66.0 &65.4 &64.2 &\textbf{62.3} &57.0 &50.1 &\emph{(10,1.0)} &51.7 &59.7 &63.6 &64.1 &\textbf{62.0} &\textbf{59.9} &57.3\\
\hline
\emph{4.0} &62.1 &63.6 &63.3 &62.5 &61.3 &\textbf{57.9} &\textbf{52.3} &\emph{(11,1.0)} &47.5 &55.5 &61.0 &61.9 &60.6 &59.3 &57.2\\
\hline
\emph{5.0} &58.9 &61.4 &61.1 &60.4 &59.6 &56.6 &52.1 &\emph{(12,1.0)} &44.0 &52.9 &58.9 &60.4 &59.5 &58.6 &\textbf{57.4}\\
\shhline[2pt]
$(l,a)$ &\emph{(0,0)} &\emph{(15,15)} &\emph{(20,20)} &\emph{(25,25)} &\emph{(30,30)} &\emph{(35,35)} &\emph{(40,40)} &$e$ &\emph{1} &\emph{1.75} &\emph{2} &\emph{2.25} &\emph{2.5} &\emph{2.75} &\emph{3}\\
\shhline[1pt]
\emph{(0,0)} &\textbf{71.7} &37.0 &27.6 &20.2 &15.4 &12.5 &10.4 &\emph{1} &\textbf{71.7} &22.1 &15.8 &11.4 &8.3 &5.9 &4.5\\
\hline
\emph{(15,15)} &56.5 &\textbf{63.8} &60.4 &51.6 &40.2 &30.8 &22.8 &\emph{1.75} &56.7 &\textbf{59.1} &55.5 &50.3 &43.8 &36.3 &29.2\\
\hline
\emph{(20,20)} &53.9 &62.3 &\textbf{61.6} &57.9 &51.3 &41.6 &32.3 &\emph{2} &52.4 &58.4 &\textbf{57.1} &53.8 &49.8 &44.7 &38.6\\
\hline
\emph{(25,25)} &48.4 &57.8 &59.3 &\textbf{59.1} &56.0 &51.1 &44.9 &\emph{2.25} &46.9 &56.6 &55.8 &\textbf{54.3} &\textbf{51.5} &48.3 &44.1\\
\hline
\emph{(30,30)} &44.5 &52.1 &54.8 &56.7 &\textbf{56.4} &53.9 &50.1 &\emph{2.5} &40.5 &53.5 &53.4 &53.0 &51.2 &49.0 &46.1\\
\hline
\emph{(35,35)} &39.4 &44.9 &47.9 &51.7 &54.2 &\textbf{54.0} &\textbf{51.9} &\emph{2.75} &39.0 &50.2 &51.0 &51.0 &50.1 &\textbf{49.1} &\textbf{46.7}\\
\hline
\emph{(40,40)} &34.0 &38.8 &42.0 &46.2 &50.0 &51.4 &51.4 &\emph{3} &33.9 &46.5 &48.2 &48.4 &48.3 &47.6 &46.3\\
\shhline[2pt]
\end{tabular}%
}
\end{table*}

\subsection{Results on Synthetic Images -- Mixed Degradation Levels}

In practice, we may not know exactly the degradation levels of real images, and it can be difficult to guarantee that the degradation levels of the testing images match those of the training images. Therefore, it is important to study the case where training images mix a wide range of image degradation levels.

For each kind of degradation, we combine the training images of all different degradation levels to generate a mixed training set for CNN training. Then, we test the CNN classifiers on testing images of the same degradation kind at each degradation level. Results are shown in Table~\ref{tab:mix-accuracy}: the four subtables from top to bottom are classification accuracy (\%) of hazy images, underwater images, motion-blurred images and fish-eye images by using VGGNet-16, respectively. The first, third, fifth and seventh rows indicate the degradation level of the testing set and the second, fourth, sixth and eighth rows indicate the classification accuracy (\%) on the corresponding degradation level of the testing set. 

From Table~\ref{tab:mix-accuracy}, we can see that the classification accuracies of clear images in its four subtables are 79.7\%, 80.5\%, 76.2\% and 77.2\% respectively, which are lower than the clear-image classification accuracy of 81\% shown in Table~\ref{tab:accuracy-vggnet}, where training images only contain clear images.  This indicates that the inclusion of degraded images into training may affect the classification accuracy of clear images. However, we can also see that the accuracy of each degradation kind and each degradation level in Table~\ref{tab:mix-accuracy} is usually higher than the accuracy of the corresponding degradation kind and level shown along the diagonal of the four subtables in Table~\ref{tab:accuracy-vggnet}, except for the degradation-free clear images. For example, accuracy of $\beta=4.0$ hazy images are 73.7\% when training set mixes all degradation level images, while this accuracy is only 72.3\% when training only on $\beta=4.0$ hazy images. This indicates that, if we know that the test images are degraded, we may want to include as many degraded images as possible, even of different degradation levels, into the training set.

\begin{table}[htbp]%
\renewcommand\arraystretch{1.45}
\setbox0\hbox{\verb/\documentclass/}%
\caption{Classification accuracy (\%) when VGGNet-16 CNN classifiers are trained by mixing different-level training images of the same degradation kind.}
\label{tab:mix-accuracy}
\centering
\setlength{\tabcolsep}{0.8mm}{
\begin{tabular}{!{\shvline[2pt]}c!{\shvline[1pt]}c|c|c|c|c|c|c!{\shvline[2pt]}}
\shhline[2pt]
$\beta$ &\emph{0} &\emph{1.5} &\emph{2.0} &\emph{2.5} &\emph{3.0} &\emph{4.0} &\emph{5.0} \\
\hline
Accuracy &79.7 &78.6 &78.1 &77.4 &76.4 &73.7 &69.2 \\
\shhline[2pt]
$(\tau,\omega)$ &\emph{(0,0)} &\emph{(6,0.8)} &\emph{(7,0.9)} &\emph{(8,1.0)} &\emph{(10,1.0)} &\emph{(11,1.0)} &\emph{(12,1.0)} \\
\hline
Accuracy &80.5 &80.1 &79.8 &78.9 &77.9 &76.9 &75.8 \\
\shhline[2pt]
$(l,a)$ &\emph{(0,0)} &\emph{(15,15)} &\emph{(20,20)} &\emph{(25,25)} &\emph{(30,30)} &\emph{(35,35)} &\emph{(40,40)} \\
\hline
Accuracy &76.2 &71.4 &69.4 &67.4 &65.5 &63.4 &61.3 \\
\shhline[2pt]
$e$ &\emph{1} &\emph{1.75} &\emph{2} &\emph{2.25} &\emph{2.5} &\emph{2.75} &\emph{3} \\
\hline
Accuracy &77.2 &74.2 &72.5 &70.8 &68.8 &66.8 &64.8 \\
\shhline[2pt]
\end{tabular}%
}
\end{table}

\subsection{Results on Real Hazy Images}

We conduct experiments in Haze-20 and HazeClear-20 datasets using VGGNet-16 and AlexNet, respectively. The experimental results are shown in Table~\ref{tab:haze-20-accuracy}. The first column indicates the kind of the training images and the second row indicates the kind of test images. where ``Combine" indicates the combination of the haze and clear training images for training. We can see that when we train and test on clear images, the accuracy can get up to 98.0\% using VGGNet-16. However, when we train and test on real hazy images, the accuracy drops to only 81.2\% using VGGNet-16. When the training set mixes haze and clear images, the test accuracy on clear images is 97.7\% and on haze image is 76.7\%. Training and testing in the same level images is the best way to achieve the better performance.


\begin{table}[htbp]%
\renewcommand\arraystretch{1.5}
\setbox0\hbox{\verb/\documentclass/}%
\caption{Classification accuracy (\%) on Haze-20 and HazeClear-20 datasets using VGGNet and AlexNet, respectively.}\label{tab:haze-20-accuracy}
\centering
\setlength{\tabcolsep}{1.5mm}{
\begin{tabular}{!{\shvline[2pt]}c!{\shvline[1pt]}c|c|c|c!{\shvline[2pt]}}
\shhline[2pt]
 &\multicolumn{2}{c|}{VGGNet-16} &\multicolumn{2}{c!{\shvline[2pt]}}{AlexNet}\\
\hline
 &HazeClear-20 &Haze-20 &HazeClear-20 &Haze-20 \\
\shhline[1pt]
HazeClear-20 &\textbf{98.0} &62.3 &\textbf{97.2} &49.9\\
\hline
Haze-20  &88.7 &\textbf{81.2} &82.0 &\textbf{75.4}\\
\hline
Combine &97.7 &76.7 &96.5 &68.1\\
\shhline[2pt]
\end{tabular}
}
\end{table}
 
\subsection{Hidden-Layer Features}

\begin{figure}[!t]
\centering
\includegraphics[width=3.3in]{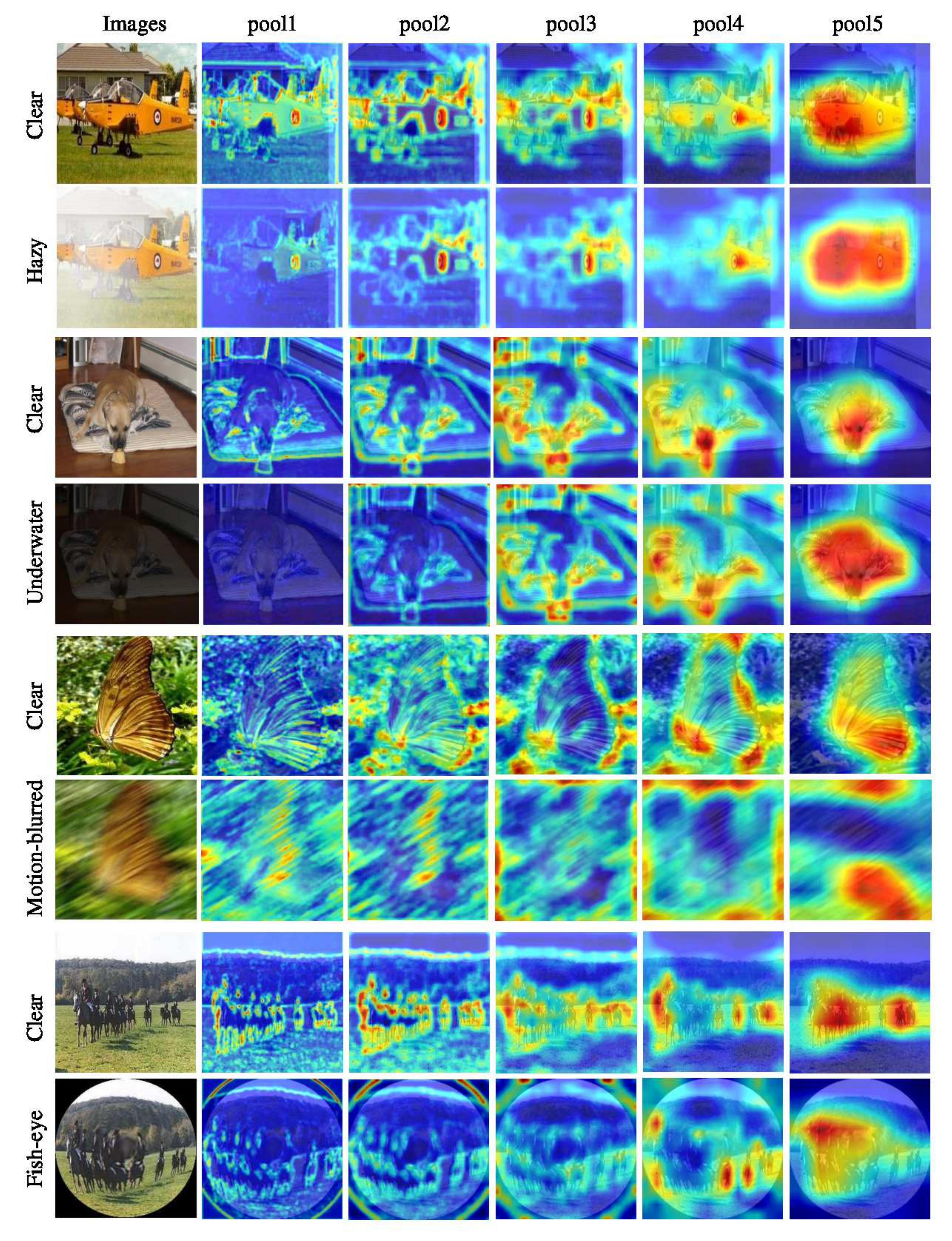}
\caption{Activations of hidden layers of CNN on image classification. From left to right are input images, and the activations at  $pool_1$, $pool_2$, $pool_3$, $pool_4$, and $pool_5$ layers, respectively.}
\label{fig:visualization}
\end{figure}

We can scrutinize the features extracted at each hidden layer to analyze the possible reasons that cause the performance drop in degraded image classification.  For an input image with size $H\times W$, the activations of a convolution layer is formulated as an order-3 tensor with $H\times W\times D$ elements, where $D$ is the number of channels. The term ``activations" is a feature map of all the channels in a convolution layer. The activations in classifying several samples images are displayed in Fig.~\ref{fig:visualization}, where five columns on the right are the activations of the max-pooling of the first, second, third, fourth, and last convolution layer in VGGNet-16, respectively. They are labeled as $pool_1$,  $pool_2$,  $pool_3$,  $pool_4$,  and $pool_5$ respectively in Fig.~\ref{fig:visualization}. For better visual effects, we resize those activations using the bi-cubic interpolation, such that they have the same size as the input image. 

We can see that, compared to clear images, the activations of ``$pool_1$" and ``$pool_2$" of the degraded images are not as discriminative. As we all know, low layers of CNN reflect color, texture, and other low-level image features. The low quality of these low-level features can very negatively affect the extraction of good-quality high-level features in later layers, leading to decreased image-classification accuracies.  Besides, from the high-level features of degraded image, we can see that the salient region is not accurately localized. For example, for the synthesized underwater image in Fig.~\ref{fig:visualization}, the degradation leads to too many salient regions which make the dog difficult to be localized. As a result, on such a degraded image, the dog is mistakenly recognized as a raccoon. Similarly, the butterfly in the motion-blurred image in Fig.~\ref{fig:visualization} is mistakenly recognized as a mushroom. Distortion changes the shape and appearance of objects, leading to incorrect features in CNN layers and finally errors in image classification. For example, in the fish-eye image in Fig.~\ref{fig:visualization}, the distortion leads to the incorrect recognition of the horse as a rifle.  

\subsection{Does Degradation-Removal Pre-Processing Help?}

As discussed earlier, for many kinds of image degradations, many researches have been conducted to remove/reduce the degradation to restore the underlying clear images~\cite{he2011single,zhu2015fast,berman2016non,cai2016dehazenet,ren2016single,
li2017aod,li2015nighttime,zhang2017fast,yau2013underwater,kim2016accurate,
sun2015learning,kannala2006generic,Hughes2010Equidistant,ying2006fisheye,li2018fisheye}. One interesting problem is whether we can get better classification accuracy by training the CNN classifiers on the clear images, and testing on the restored test images after the degradation removal. In this section, on the synthetic data we pick the haze-removal algorithm developed in~\cite{he2011single}, the deblurring algorithm developed in~\cite{Fergus2006Removing}, and distortion correction algorithm developed in~\cite{kannala2006generic} to remove the corresponding 
degradations in the test images and then run the CNN-based image classification. Results are shown in Table~\ref{tab:DE}, which contains three subtables, separated by thick lines, for hazy images (top), motion-blurred images (middle) and fish-eye images (bottom), respectively. For each subtable, the row of ``w/o DM" indicates the results of training on clear images and testing on degraded images without the degradation removal, the row of ``w DM" indicates the results of training on clear images and testing on degraded images with the degradation removal, and the row of ``Diag." indicates the results of training and testing on the images of the same degradation level, copied here from the diagonal of the corresponding subtables in Table~\ref{tab:accuracy-vggnet}.
 
We can see that, if we train the CNN model on clear images, a pre-processing of degradation removal can sometimes help improve the classification accuracy, especially for hazy and motion-blurred images. But the degradation removal could never lead to a classification accuracy close to the level of training and testing both on original clear images. Comparing the rows of ``w DM" and ``Diag.", we can also see that, the accuracy resulting from training and testing on the same degradation level is much higher than the accuracy resulting from training on clear images and testing on the images recovered from degradation removal. This shows that the degradation-removal algorithms may transform the images to be more visually pleasant to human eyes, but may not help much for CNN-based classification: for degraded images, training and testing directly without degradation removal actually produces the best accuracy. We believe this is reasonable since the degradation-removal algorithms do not introduce any new information to the CNN-based classification.
 
\begin{table}[htbp]%
\renewcommand\arraystretch{1.5}
\setbox0\hbox{\verb/\documentclass/}%
\caption{Classification accuracy (\%) when training on clear images and testing on degraded images with and without the degradation removal. The three subtables, separated by thick lines, are for hazy images (top), motion-blurred images (middle) and fish-eye images (bottom), respectively.} \label{tab:DE}
\centering
\setlength{\tabcolsep}{1mm}{
\begin{tabular}{!{\shvline[2pt]}c!{\shvline[1pt]}c|c|c|c|c|c|c!{\shvline[2pt]}}
\shhline[2pt]
$\beta$ &\emph{0} &\emph{1.5} &\emph{2.0} &\emph{2.5} &\emph{3.0} &\emph{4.0} &\emph{5.0}\\
\shhline[1pt]
w/o DM &81.0 &74.6 &68.7 &61.3 &54.0 &41.8 &33.0\\
\hline
w DM &81.0 &75.7 &74.3 &71.8 &67.1 &55.0 &43.0\\
\hline
  Diag.   &\textbf{81.0} &\textbf{78.6} &\textbf{77.7} &\textbf{76.6} &\textbf{75.3} &\textbf{72.3} &\textbf{67.8}\\
\shhline[2pt]
$(l,a)$ &\emph{(0,0)} &\emph{(15,15)} &\emph{(20,20)} &\emph{(25,25)} &\emph{(30,30)} &\emph{(35,35)} &\emph{(40,40)}\\
\shhline[1pt]
w/o DM &81.0 &45.1 &31.2 &22.6 &16.6 &13.3 &11.2\\
\hline
w DM &81.0 &50.3 &35.0 &25.4 &18.4 &14.4 &12.0\\
\hline
 Diag.    &\textbf{81.0} &\textbf{72.6} &\textbf{69.9} &\textbf{66.9} &\textbf{64.2} &\textbf{61.6} &\textbf{59.0}\\
\shhline[2pt]
$e$ &\emph{1} &\emph{1.75} &\emph{2} &\emph{2.25} &\emph{2.5} &\emph{2.75} &\emph{3}\\
\shhline[1pt]
w/o DM &81.0 &25.4 &17.6 &12.4 &9.4 &7.0 &5.1\\
\hline
w DM &81.0 &20.0 &14.0 &10.0 &7.3 &5.8 &5.2 \\
\hline
Diag. &\textbf{81.0} &\textbf{73.7} &\textbf{71.0} &\textbf{69.2} &\textbf{66.9} &\textbf{64.3} &\textbf{62.7}\\
\shhline[2pt]
\end{tabular}%
}
\end{table}

\section{Conclusion} \label{sec:conclusion}
In this paper, we conducted an empirical study to explore the effect of four kinds of image degradations to the performance of CNN-based image classification. For facilitating the quantitative evaluation, we proposed to synthesize a large number of images for training and testing. We considered the synthesis of hazy images, underwater images, motion-blurred images and fish-eye images, each with seven degradation levels, as well as collection of real hazy images from Internet. We found that the image classification performance does drop significantly when the image is degraded, especially when the training images can not well reflect the degradation levels of the test images. By visualizing the activations of hidden layers of the CNN classifiers, we found that many important low level features were not well discerned in early layers, which might be a key factor for the dropped classification accuracy. We also found that the existing algorithms for removing various kinds of degradations could not be used to improve much the CNN-based classification performance. We hope this study can draw more interests from the community to work on degraded image classification, that can benefit many important application domains such as  autonomous driving, underwater robotics, video surveillance, and wearable cameras.


%



\ifCLASSOPTIONcaptionsoff
  \newpage
\fi



%

\bibliographystyle{IEEEtran}
\bibliography{ieeetran}

\end{document}